\documentclass{article}

\usepackage[final]{corl_2019} 
\usepackage{acronym}

\usepackage{algorithm}
\usepackage{algorithmic}
\usepackage{hyperref}
\usepackage{mathtools}
\usepackage{amsmath}
\usepackage{amssymb}
\usepackage{float}
\usepackage{siunitx}
\usepackage{adjustbox}
\usepackage{array}
\usepackage{caption} 
\usepackage{booktabs}
\usepackage[table, xcdraw]{xcolor}
\usepackage{subcaption}
\usepackage[para]{footmisc}
\usepackage{multicol}
\usepackage{moresize}
\newcolumntype{R}[2]{%
    >{\adjustbox{angle=#1,lap=\width-(#2)}\bgroup}%
    l%
    <{\egroup}%
}

\usepackage[utf8]{inputenc} 
\usepackage[T1]{fontenc}    

\title{
Navigation Agents for the Visually Impaired:\\ A Sidewalk Simulator and Experiments 
}

\author{
  \textbf{Martin Weiss$^{\ddagger}$
  Simon Chamorro$^{\ast}$ 
  Roger Girgis$^{\diamond}$ 
  Margaux Luck$^{\ddagger}$ } \\
  \textbf{Samira E. Kahou$^{\triangleright}$ 
  Joseph P. Cohen$^{\ddagger}$
  Derek Nowrouzezahrai$^{\triangleright}$ }\\
  \textbf{Doina Precup$^{\triangleright}$
  Florian Golemo$^{\ddagger, \S}$ 
  Chris Pal$^{\diamond, \S}$}   \And   \vspace{-20pt}\\
  \footnotesize{$^{\ddagger}$ Université de Montréal,
                $^\ast$Université de Sherbrooke, $^{\diamond}$Polytechnique Montréal,}\\
   \footnotesize{$^{\triangleright}$McGill University,
                $^{\S}$ElementAI, Correspondence: \href{mailto:martin.clyde.weiss@gmail.com}{martin.clyde.weiss@gmail.com}
  }\\
  \footnotesize{All authors are members of Mila, Quebec Artificial Intelligence Institute}}

\begin{document}

\acrodef{FOV}{Field of View}
\acrodef{FPS}{Frames Per Second}
\acrodef{OCR}{Optical Character Recognition}
\acrodef{RL}{Reinforcement Learning}
\acrodef{PPO}{Proximal Policy Optimization}
\acrodef{SLAM}{Simultaneous Localization and Mapping}
\acrodef{DQN}{Deep Q-Network}
\acrodef{DRL}{Deep Reinforcement Learning}

\maketitle

\vspace{-10pt}
\begin{abstract}
Millions of blind and visually-impaired (BVI) people navigate urban environments everyday, using smartphones for high-level path-planning and white canes or guide dogs for local information.
However, many BVI people still struggle to travel to new places. 
In our endeavour to create a navigation assistant for the BVI, we found that existing \ac{RL} environments were unsuitable for the task.
This work introduces SEVN, a sidewalk simulation environment and a neural network-based approach to creating a navigation agent.
SEVN contains panoramic images with labels for house numbers, doors, and street name signs, and formulations for several navigation tasks.
We study the performance of an \ac{RL} algorithm (PPO) in this setting. Our policy model fuses multi-modal observations in the form of variable resolution images, visible text, and simulated GPS data to navigate to a goal door. 
We hope that this dataset, simulator, and experimental results will provide a foundation for further research into the creation of agents that can assist members of the BVI community with outdoor navigation.

\end{abstract}

\keywords{outdoor navigation, computer vision, reinforcement learning, dataset} 


\section{Introduction}  
There are currently around 250 million blind and visually-impaired (BVI) people \cite{bourne2017magnitude}, many of whom struggle with common tasks such as sorting mail and medications, reading, and separating laundry due to their impairment.
To understand the problems BVI people encounter, how they are solved today, and where progress is being made, we engaged in a design process in collaboration with BVI people, specialists in orientation and mobility, and developers (detailed in supplementary section \ref{user-centric-design}).
A notable challenge for BVI people is finding specific locations in urban environments.
Navigating the last meters from the sidewalk to a new building entrance unassisted by humans remains an open problem.
Today, BVI people use GPS-enabled smartphones for long-distance route planning but often must rely on human assistance for short-range tasks. Remote support by humans is offered by applications like Aira or BeMyEyes, but may be costly or slow, and may not preserve privacy.
Fully-automated mobile applications like SeeingAI, EnvisionAI, and SuperSense 
can identify objects, extract text, or describe scenes to improve BVI autonomy, but are not designed for navigation.

We wanted to design an automated system for BVI pedestrian navigation which combined visuospatial perception with real-time decision-making.
With that in mind, we analyzed visually realistic simulators for autonomous driving \cite{CarlaDosovitskiy2018,airsim2017fsr,Lgsvl} only to find they lack realistic buildings and scene text. 
Real-world image datasets containing these features are almost exclusively captured from vehicle-mounted cameras \cite{Bdd100kYu2018,CityscapesCordts2016}. 
And though specialized datasets for street crossing and pedestrian localization exist, they are static and contain limited ranges of imagery \cite{talkthewalk2018Devries,FreiburgRadwan2018}.
To the best of our knowledge, none of these datasets can support the development of robust sidewalk navigation models.

\begin{figure}[tb]
\centering
  \centering
  \includegraphics[width=1.0\linewidth]{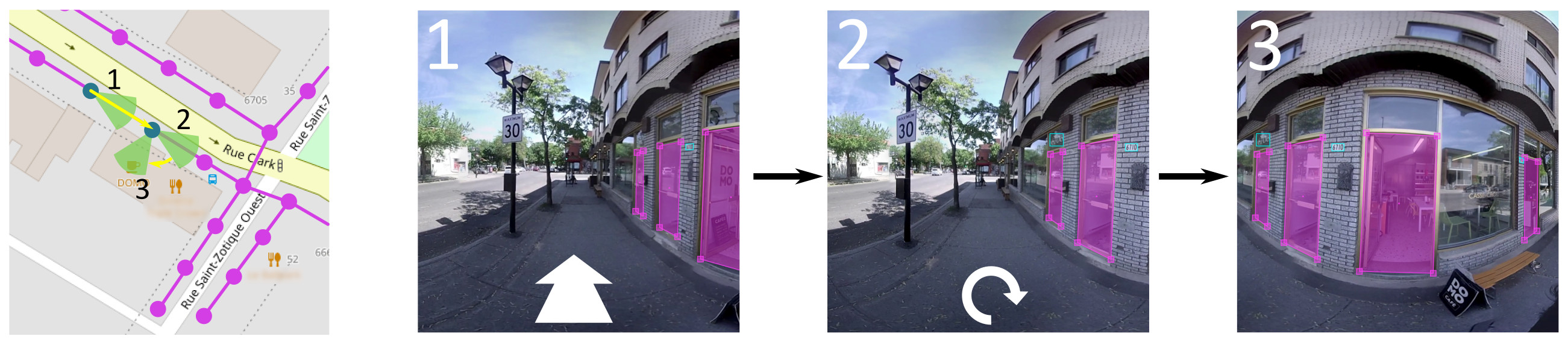}
  \captionof{figure}{\textbf{Illustrated example of an agent trajectory.}
  Left) An overhead view of the trajectory. 1) The agent starts on the sidewalk with the goal door in view but out of reach; the agent moves forward. 2) The goal door is beside the agent; the agent turns right. 3) The correct door is visible and fully contained within the frame; the task is complete.
  }
  \label{fig:twitter}
\end{figure}

In this work, we introduce a novel simulated environment for sidewalk navigation "SEVN" (read "seven"), that supports the development of assistive technology for BVI pedestrian navigation.
SEVN contains $4,988$ high resolution panoramas with $3,259$ labels on house numbers, doors, and street signs (see Figure \ref{fig:twitter} for an example).
These panoramas are organized as an undirected graph where each node contains geo-registered spatial coordinates and a \ang{360} image. 
Additionally, we define several pedestrian navigation tasks that may be completed using multi-modal observations in the form of variable resolution images, extracted visible text, and simulated GPS data.
Our experiments focus on learning navigation policies that assume access to ground truth text labels, and in this setting our multimodal fusion model demonstrates strong performance on a street segment navigation task.
We hope that the release of this paper, dataset, and code-base will spark further development of agents that can support members of the BVI community with outdoor navigation\footnote{\url{https://mweiss17.github.io/SEVN}}. 

The primary contributions of this paper are:
\begin{itemize}
  \item A benchmark dataset containing panoramic images from over six kilometers of sidewalks in Little Italy, Montreal, with annotated house numbers, street signs, and doors\footnote{\url{https://github.com/mweiss17/SEVN-data}}.
  \item An OpenAI Gym-compatible environment \cite{1606.01540} for \ac{RL} agents with multi-resolution real-world imagery, visible text, simulated GPS, and several task settings\footnote{\url{https://github.com/mweiss17/SEVN}}.
  \item A novel neural architecture for RL trained with \ac{PPO}\cite{DBLP:journals/corr/SchulmanWDRK17} to fuse images, GPS, and scene text for navigation, with results and ablations \footnote{\url{https://github.com/mweiss17/SEVN-model}}.
\end{itemize}

\section{Related Work}
\textbf{Assistive technology for the BVI} has been an active area of research for many years, with most work being designed in collaboration with the BVI or featuring extensive user studies. 
A good example of this practice is the VizWiz dataset \cite{Bigham:2010:VNR:1866029.1866080} which consists of 31,000 questions and images generated by BVI people, with corresponding answers from sighted people.
While useful for benchmarking VQA models for the BVI, the VizWiz dataset focuses more on object identification, reading, and description than navigation \cite{Brady:2013:VCE:2470654.2481291}. 
Recent works on BVI navigation tend to focus on the problems of intersection crossing \cite{Diaz_2017_ICCV} and obstacle detection \cite{radwan2018multimodal}.
In the context of sidewalk navigation, options like the BeAware app
propose a combination of beacons planted in the environment and bluetooth communication to assist the BVI.
To the best of our knowledge, the problem of sidewalk navigation to a specific location seems to have limited prior work.

\textbf{Vision and language navigation} tasks generally require agents to act in the environment to achieve goals specified by natural language.
This setting at the intersection of computer vision, natural language processing, and reinforcement learning has generated many tasks and a common framework for evaluation of embodied navigation agents \cite{DBLP:journals/corr/abs-1807-06757}.
The task defined in Vision and Language Navigation \cite{mattersim} provides agents with 21,567 human-generated visually-grounded natural language instructions for reaching target locations in the Matterport3D environment \cite{Matterport3D}, an environment which consists of 10,800 panoramic views from 90 building-scale scenes connected by undirected spatial graphs. 
The success of \cite{mattersim} has also motivated the development of systems to generate natural language instructions for navigation \cite{fried2018speaker}.
Many environments focus on navigating apartment interiors, with continuous \cite{xiazamirhe2018gibsonenv} or discrete \cite{Matterport3D} action spaces and varying levels of interactivity \cite{DBLP:journals/corr/abs-1712-05474}, while relatively few have investigated vision and language tasks in outdoor settings.
 

\textbf{Outdoor environments} often base their work on panoramic imagery and connectivity graphs from Google Street View. 
These environments support tasks that range from navigating to the vicinity of arbitrary coordinates \cite{mirowski} and certain types of locations (e.g. gas stations, high schools) \cite{DBLP:journals/corr/BrahmbhattH17}, to following instructions \cite{DBLP:journals/corr/abs-1903-00401}.
However, the sparsity of these nodes (averaging 10 meters between connected nodes) and the vehicle-mounted perspective imagery makes these environments unsuitable for pedestrian navigation.
Furthermore, such environments do not provide the type of labelled information that is necessary to construct pedestrian navigation tasks (e.g., door numbers, street signs, door annotations).
Nor is it possible to provide dense annotations in the StreetLearn Google Street View data \cite{DBLP:journals/corr/abs-1903-01292} because the resolution of that imagery is too low (1664 x 832 pixels).
As outlined in Section \ref{sec:env}, SEVN provides the higher resolution imagery that is necessary for agents that find and reason about scene text in order to navigate to specified doors.

\section{A Sidewalk Environment for Visual Navigation (SEVN)}
\label{sec:env}
This section introduces SEVN, a visually realistic \ac{DRL} simulator for pedestrian navigation in a typical urban neighborhood. 
We first describe the process by which we captured and annotated the image data in SEVN before discussing the simulator's interface.

\textbf{The Data} was first captured as \ang{360} video using a Vuze+ camera attached to the top of a monopod held slightly above the operator's head. 
The Vuze+ has four synchronized stereo cameras.
Each stereo camera is composed of two image sensors with fisheye lenses that each capture full high definition video (1920x1080) at $30$ \ac{FPS}.
We used the VuzeVR Studio software to stitch together the raw footage from each camera to obtain a \ang{360} stabilized video, from which we extracted 3840 $\times$ 1920 pixel equirectangular projections. We then crop and remove the top and bottom sixth of these panoramas resulting in a 3840 $\times$ 1280 image which contains about 3.5 times as many pixels as those in StreetLearn \cite{DBLP:journals/corr/abs-1903-01292}. 

\begin{figure}[ht]
\centering
\includegraphics[height=70pt]{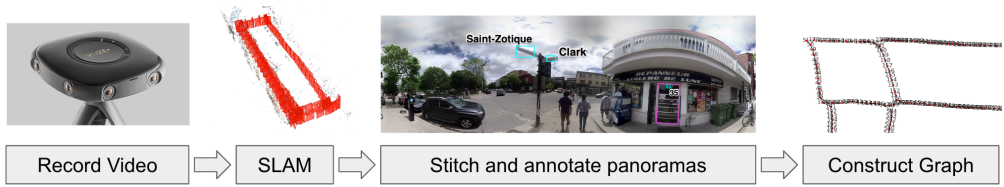}
\caption{
\textbf{Data processing pipeline}. We captured footage using a Vuze+ array of 4 stereo cameras. We then localized this footage with ORB-SLAM2 \cite{MurArtal2017ORBSLAM2AO} yielding a spatial graph of images and positions. We sparsified the graph, stitched the remaining images into \ang{360} panoramas using VuzeVR Studio, and hand-annotated them.
}
\label{fig:pipeline}
\end{figure}

We used ORB-SLAM2 \cite{MurArtal2017ORBSLAM2AO}, a \ac{SLAM} pipeline, to geo-localize the footage. 
As input to ORB-SLAM2, we provided an undistorted view of the left front facing camera from the Vuze+ and obtained a camera pose for each localized frame (3-D coordinate and quaternion). 
From these camera poses, we created a 2-D graph with node connectivity determined by spatial location. 
The recording framerate (30 Hz) resulted in very little distance between nodes - so we sparsified the graph to 1 node per meter.
Finally, we manually curated the connectivity of the graph and location of the nodes. Figure \ref{fig:pipeline} summarizes the data processing pipeline. 

Our dataset is split into a Large and Small dataset, where Small is a block containing several urban zones (e.g. residential, commercial) bounded by St. Laurent, St. Zotique, Baubien, and Clark. Montreal's municipal zoning regulations are described in supplementary section \ref{zoning}. Figure \ref{fig:map} shows the full graph with these splits overlaid on a map. Each split contains street segments and intersections. As part of this work, we released the code to capture, SLAM, and filter the data\footnote{\url{https://github.com/mweiss17/SEVN-data}}.

\begin{figure}[htb]
\centering
  \includegraphics[width=.8\linewidth]{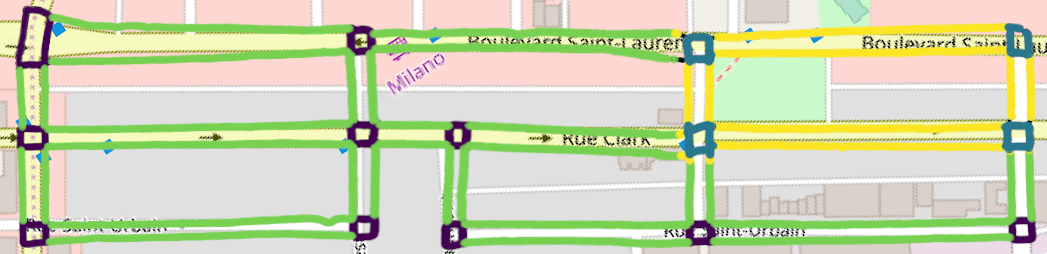}
  \captionof{figure}{\textbf{SEVN Spatial Graph} super-imposed on an OpenStreetMap Carto (Standard) view of Little Italy, Montreal. The dataset is split into two sets: Large and Small. The Large dataset street segments shown in green and the intersections are shown as purple. Street segments in the Small dataset are shown in yellow and the intersections are shown in teal. Visualizations of the goal locations and municipal zoning are contained in the supplemental material, sections \ref{goal-locs} and \ref{zoning}.}
  \label{fig:map}
\end{figure}

\textbf{The Annotations} we provide on the full-resolution panoramas are bounding boxes around all visible street name signs and house numbers, and include ground truth text annotations for both. 
Doors with clearly associated house numbers are annotated with polygons. 
To create the annotations, we used the "RectLabel" software. 
The annotated panoramas are publicly available with blurred faces and licence plates\footnote{\url{https://github.com/mweiss17/SEVN\#dataset}}. 
These annotations and privacy blurs were added and verified by two authors.
Table \ref{tab:env-data} shows some global statistics about the dataset, the splits, and the annotations.

\label{sec:env-ann}
\begin{table}[htb]
\begin{center}
\begin{tabular}{@{}lllllll@{}}
\toprule
\textbf{Dataset} & \textbf{\# Images} & \textbf{\# Streets} & \textbf{Length (m)} & \textbf{\# Doors} & \textbf{\# Signs} & \textbf{\# House Numbers} \\ \midrule
Large    & 3,831     & 8          & 5,000      & 1,017(355) & 167 (11)         & 1,307 (367)           \\
Small & 1,157     & 4          & 1,300        & 286 (116)   & 46  (3)         & 436  (119)           \\ 
Total    & 4,988     & 8          & 6,300      & 1,303 (470) & 213 (11)         & 1,743 (485)           \\
\bottomrule
\end{tabular}
\end{center}
\caption{
\textbf{Dataset statistics and number of annotations}. We report the total amount of house number, door, and street name sign annotations with the corresponding number of unique physical objects that were annotated in parentheses. Inside the parentheses of the "\# Signs" column, we report the unique number of street names visible within the dataset, not the unique number of street name signs. The "\# Streets" column shows the number of unique streets where we captured data.
}
\label{tab:env-data}
\end{table}

\textbf{The SEVN Simulator} is based on the OpenAI Gym \cite{1606.01540} environment.
We chose a similar action space to Learning to Navigate in Cities without a Map \cite{mirowski} with slight left and right turns ($\pm$\ang{22.5}), sharp left and right turns ($\pm$\ang{67.5}), and a forward action that transitions to the neighboring node nearest the agent's current heading.
If there is no node within $\pm$ \ang{45} of the agent's current heading, then we do not change the agent's position.
We also propose two other actions which are used in a subset of the tasks proposed in section \ref{sec:tsk}: read and done. 
The read action is only used in the CostlyTxt task where the agent incurs a small negative reward in order to access the scene text; the done action is only used in the Sparse task to terminate the trajectory.
The agent observes a \ang{140} normalized image cropped from a low-resolution ($224\times84$ px) version of the panorama, created during a pre-processing step from the high-resolution panoramas ($3840\times1280$ px). 
At this low resolution, most text becomes illegible.
Therefore, at each timestep we check if any text which was labelled in the full scale panorama is fully contained in the agent's \ac{FOV}; we encode these labels; and we pass them as observations to the agent (see section \ref{sec:tsk-modalities}).
An instance of the simulator running with low-resolution imagery can be run at 400-800 \ac{FPS} on a machine with 2 CPU cores and 2 GB of RAM.

\textbf{The Oracle} was implemented to determine the shortest navigable trajectory between any two poses. 
A pose is defined as the combination of an agent heading (discritized into \ang{22.5} wedges) and a position (restricted to the set of localized panoramas).
Our panorama graph is implemented in NetworkX \cite{SciPyProceedings_11}, which provides a function to find the shortest path between two nodes ($n_1, n_2)$ in graph $G$, i.e., $nx.shortest\_path(G, source=n_1, target=n_2)$.
For each node we calculated the most efficient way to turn in order to face the next node before emitting a 'forward' action. 
Once the agent is located in the goal node, we check the direction of the target door and turn until it is entirely contained within the agent's \ac{FOV}.


\section{Reinforcement Learning, Rewards \& Tasks}
\label{sec:tsk}

In this work we are interested in exploring the use of Reinforcement Learning (RL) techniques for learning agents that have the goal of assisting with navigation tasks.
There are many navigation policies which can be learned to assist the BVI, each with its own strengths and weaknesses.
This section proposes several reward structures and tasks for learning agents within an RL framework that encourage agents to learn different navigation policies.
These are summarized in Table \ref{tab:tsk-envs}.
\subsection{Proposed Rewards}
\label{propreward}
\textbf{Dense rewards} are provided to the agent at each time-step. We provide a $+1$ reward for each transition taking the agent closer to the goal, and a $-1$ for each transition that takes the agent further from the goal.
For turn actions, we provide a $-.2$ for turns away from the direction which enables a correct transition, and we provide a $+.1$ reward for each turn towards this direction, unless the agent has already inhabited a pose nearer the correct direction in which case the action returns a 0 reward.
Within a goal node the correct direction is a heading which fully contains the target door polygon.

\textbf{Costly Read} is a setting where a small negative reward, $-.2$, is provided to the agent after taking the "read" action, returning the labelled text within the \ac{FOV}.
This reward would encourage the agent to only take this action when it provides improved localization. 
In the context of an expensive scene text recognition model, a policy trained with this reward could be more computationally efficient.

\textbf{Multi-Goal reward} is a setting wherein the agent receives a positive reward, $+1$, for each house number it "sees" during an episode. In this setup, the agent is encouraged to navigate towards regions with many visible house numbers. This behaviour could be useful in assistive navigation systems, helping the BVI person to find house numbers to augment their own navigation ability.

\textbf{Sparse rewards} not only require the agent to complete the entire task before receiving a reward, but also require the agent to emit a "done" action with the target door fully in view. This task is quite challenging, but successful models should generalize better than those trained with dense rewards.

\subsection{Tasks}
In tasks 1-7, we require that the agent occupy the  node closest to the goal door.  
We identified the nearest panorama to a goal address through a proxy metric: the polygon with the largest door area for a given house number.
The success condition also requires the target door to be entirely contained in the agent’s field of view. 
For all tasks, we first select a valid terminal state by uniformly sampling over doors with addresses. 
Next, we identify the street segment which contains this address and uniformly sample the agent’s start node and direction from this segment.

\textbf{Tasks 1 to 4 } represent an ablation study wherein the agent is trained with different combinations of sensor modalities to determine their relative contributions in a dense reward setting. This task investigates the small-scale sidewalk navigation problem, with trajectories terminating when the agent has navigated to the goal node and turned so that the goal door is fully within view. 

\textbf{Task 5} is equivalent to task 1, but the start and end poses are not restricted to the same street segment. Instead, the agent must navigate through an intersection to a goal on another street segment. 

\textbf{Tasks 6, 7, and 8} are equivalent to task 1, except that the agent is trained with the costly read reward structure, the sparse reward structure and the multi-goal reward structure, respectively.

\begin{table}[htb]
\begin{center}
\begin{tabular}{@{}llllllllll@{}}
\toprule
   &                & \multicolumn{3}{c}{\cellcolor[HTML]{EFEFEF}\underline{\textbf{Observations}}}                                    & \multicolumn{3}{c}{\underline{\textbf{Rewards}}}                                                                                                                                                    \\ 
\textbf{ID} & \textbf{Task Name}           & \cellcolor[HTML]{EFEFEF}\textbf{Img} & \cellcolor[HTML]{EFEFEF}\textbf{GPS} & \cellcolor[HTML]{EFEFEF}\textbf{Txt} & \textbf{Dense} & \textbf{Costly Read}  & \textbf{Sparse} & \textbf{Multi-Goal} \\\midrule
1  & AllObs     & \cellcolor[HTML]{EFEFEF}\checkmark   & \cellcolor[HTML]{EFEFEF}\checkmark   & \cellcolor[HTML]{EFEFEF}\checkmark  & \checkmark & $\cdot$ & $\cdot$ & $\cdot$ \\
2  & NoImg    & \cellcolor[HTML]{EFEFEF}$\cdot$   & \cellcolor[HTML]{EFEFEF}\checkmark   & \cellcolor[HTML]{EFEFEF}\checkmark  & \checkmark & $\cdot$ &  $\cdot$ & $\cdot$ \\
3  & NoGPS   & \cellcolor[HTML]{EFEFEF}\checkmark   & \cellcolor[HTML]{EFEFEF}$\cdot$   & \cellcolor[HTML]{EFEFEF}\checkmark  & \checkmark & $\cdot$ & $\cdot$ & $\cdot$ \\
4  & ImgOnly  & \cellcolor[HTML]{EFEFEF}\checkmark   & \cellcolor[HTML]{EFEFEF}$\cdot$   & \cellcolor[HTML]{EFEFEF}$\cdot$  & \checkmark & $\cdot$ & $\cdot$ & $\cdot$ \\\midrule
5 & Intersection & \cellcolor[HTML]{EFEFEF}\checkmark   & \cellcolor[HTML]{EFEFEF}\checkmark   & \cellcolor[HTML]{EFEFEF}\checkmark  & \checkmark & $\cdot$ & $\cdot$ & $\cdot$ \\
6  & CostlyTxt      & \cellcolor[HTML]{EFEFEF}\checkmark   & \cellcolor[HTML]{EFEFEF}\checkmark   & \cellcolor[HTML]{EFEFEF}\checkmark  & \checkmark & \checkmark & $\cdot$ & $\cdot$ \\
7  & Sparse         & \cellcolor[HTML]{EFEFEF}\checkmark   & \cellcolor[HTML]{EFEFEF}\checkmark   & \cellcolor[HTML]{EFEFEF}\checkmark & $\cdot$ & $\cdot$ & \checkmark & $\cdot$ \\
8  & Explorer         & \cellcolor[HTML]{EFEFEF}\checkmark   & \cellcolor[HTML]{EFEFEF}\checkmark   & \cellcolor[HTML]{EFEFEF}\checkmark  & $\cdot$ & $\cdot$ & $\cdot$ & \checkmark \\
\bottomrule
\end{tabular}
\end{center}
\caption{
\textbf{Rewards \& Tasks}. The first four tasks examine combinations of observation modalities. 
In the CostlyTxt task, the agent has access to a "read" action which yields the scene text, but imposes a small negative reward.
The Intersection tasks requires the agent to cross static intersections to find goals on other street segments. 
The Explorer task gives a reward for each unique house number the agent sees. 
The Sparse task, which is most challenging, only gives the agent a reward once it reaches its target destination and emits a "done" action to terminate the episode. }
\label{tab:tsk-envs}
\end{table}


\section{Modalities}
\label{sec:tsk-modalities}
The agent has access to three types of observation modality and a reward signal. 
An observation can contain an image, GPS, and visible text. Table \ref{tab:tsk-obs} summarizes the available modalities and formats.

The image is a forward-facing RGB image of shape (3, 84, 84) that contains a \ang{135} \ac{FOV} which depends on the agent's direction. Note that all tasks can also be run with high resolution images of shape (3, 1280, 1280).
The simulated GPS is a 2-dimensional ego-centric vector indicating the relative x and y offset from the goal in meters. 
Finally, coordinates are scaled to the range $[-1,1]$. 

We study two kinds of incidental scene text: house numbers and street name signs. 
At each timestep, we determine whether there are any house number or street name sign bounding boxes within the \ac{FOV} of the agent. 
If so, we pass them to the agent as encoded observations. 
We encode the house numbers as four one-hot vectors of length 10 (since the longest house numbers in our dataset contain 4 digits) flattened into a unique one hot vector of size (40, 1) representing integers between 0-9999. 
Next, we stack up to three of these vectors to create a vector of shape (120, 1) before passing them to the agent.
This enables the agent to see up to three house numbers simultaneously. 
Street name signs are similarly encoded, but span the 11 street names contained in the dataset.

\begin{table}[H]
\begin{center}
\begin{tabular}{@{}lll@{}}
\toprule
\textbf{Obervation Type} & \textbf{Dimensions}      & \textbf{Description}                                                                        \\ \midrule
Image (low-res)  & (3, 84, 84)     & Low-resolution RGB image observation with \ang{135} \ac{FOV}  \\
Image (high-res)  & (3, 1280, 1280)     & High-resolution RGB image observation with \ang{135} \ac{FOV}  \\
GPS Coordinates & (4, 1)          & Absolute and relative goal coordinates \\
House Number    & (40, 1)         & One-hot encoding representing integers between 0-9999                         \\
Street Name     & (11, 1) & One-hot encoding of street name signs                                     \\ \bottomrule
\end{tabular}
\end{center}
\caption{\textbf{Observation modalities and formats.} The simulator provides three observation modalities: images, GPS, and text. Images can be provided in either low resolution $(84 \times 84$ px) or high resolution $(1280 \times 1280$ px). The absolute coordinates of the goal are fed in as two floating point values $(x, y)$ scaled between -1 and 1, while the relative coordinates are computed by taking the difference between the agent's current position and the goal position. Each character in a house number is represented by a $(10, 1)$ one-hot vector; we concatenate four of these together to represent the range of house numbers found in our dataset. 
}
\label{tab:tsk-obs}
\end{table}




\section{Baseline Experiment \& Results}
\label{sec:bas}
In this section, we describe a multi-modal fusion model, trained with \ac{PPO} \cite{DBLP:journals/corr/SchulmanWDRK17}, that navigates to a target door with a 74.9\% success rate. We also ablate this model to investigate the contribution of each observation modality. See supplementary section \ref{noisy-gps} for ablations with noisy GPS sensors.

\subsection{Method \& Architecture}
\begin{figure}[ht]
\centering
\includegraphics[height=160pt]{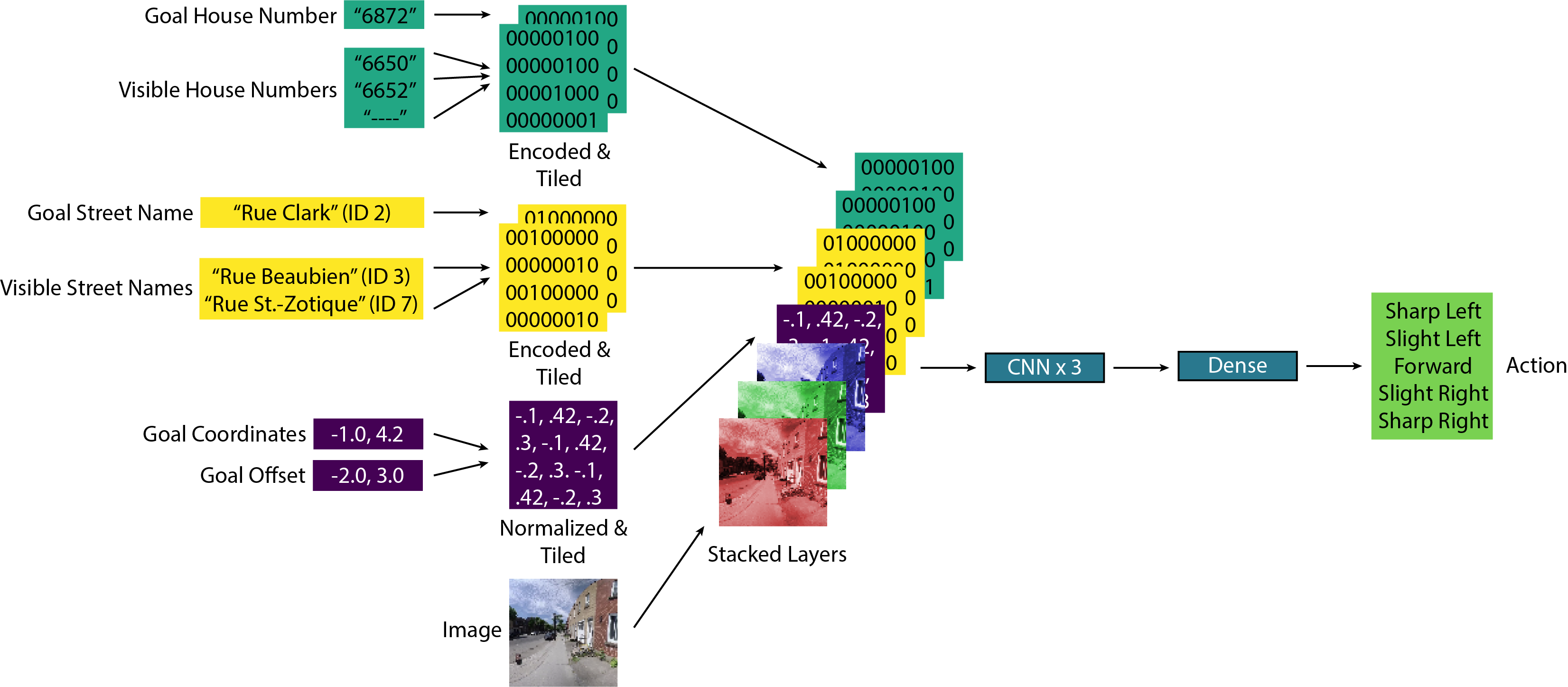}
\caption{
\textbf{Policy Network Architecture.} We show the different input modalities in both their human-readable format and their tiled format. The tiled format is then appended to the RGB image matrix creating an $(8, w, w)$ tensor, where $w$ is the width of the square input image. This tensor is then processed by 3 convolutional layers before being flattened and processed by a dense layer. Finally, this dense layer outputs a vector over the agent's action space, from which we sample the agent's action at that step.
}\label{fig:bas-net}
\end{figure}

To train our model, we chose the \ac{PPO} algorithm for its reliability without hyperparameter search\footnote{\url{https://github.com/mweiss17/SEVN-model}}. 
For our policy network (see Figure \ref{fig:bas-net}), we take each modality in its one-hot format (as described in Table \ref{tab:tsk-obs}) and tile it into a matrix of size $(1, w, w)$, where $w$ is the width of the image. We then add any necessary padding. Performing this process for each modality and appending all matrices to the input image yields a tensor of dimension $(8, w, w)$.
If a model does not have access to a modality, then that matrix (or matrices) are simply filled with zeros. 
This representation is fed through three convolutional layers with kernel sizes 32, 64, 32, followed by a dense layer of size 256, all with ReLU nonlinearities.
This network outputs a probability vector of dimension $(1, y)$, with $y$ being the dimension of the action space.
To determine the agent's next action, we sample from this distribution.
The critic network contains an additional dense network that converts the combined embeddings into a single value.
The PPO hyperparameters are mostly identical with the recommended parameters for training in Atari gym environments \cite{1606.01540}; the only difference being that we increase the learning rate (from $2.5\mathrm{e}{-4}$ to $3\mathrm{e}{-4}$) and the training duration (from $1\mathrm{e}{6}$ steps to $2\mathrm{e}{6}$). 
The full set of parameters can be found in the supplementary section \ref{ppo-params}.

\subsection{Results}
Table \ref{tab:results} reports the mean and standard deviation of policies after 2M steps of experience on the Small dataset. Results are averaged over 10 seeds. We also report the oracle and random walk performance for comparison.
In this setting, the agent has 253 timesteps to navigate to its target location, the length of the longest optimal trajectory in the environment.
A random agent successfully completed this task within 253 actions in $5.7\%$ of episodes, while the oracle completed 100 \% of tasks in 80.8 steps on average.
Two seeds were removed because our analysis indicated that they learned a degenerate solution, exploiting the environment. 

\begin{table}[ht]
\begin{center}
\resizebox{0.8\textwidth}{!}{\begin{tabular}{@{}llcccccc@{}}
\toprule
\textbf{ID} & \textbf{Task Name} & \textbf{Success Rate} & \textbf{Reward} & \textbf{Trajectory Length} \\ 
1 & AllObs  & 74.9\% ($\pm$ 7.8\%)  & 43.9 ($\pm$ 65.9)     & 113.7 ($\pm$ 89.7) &            \\
2 & NoImg   & 0.8\% ($\pm$ 0.3\%)     & -24.6 ($\pm$ 2.3)    & 251.0 ($\pm$ 22.0) &                \\
3 & NoGPS   & 58.2\% ($\pm$ 3.2\%)   & 32.8 ($\pm$ 78.3)    & 147.6 ($\pm$ 96.6) &              \\
4 & ImgOnly & 57.\%($\pm$ 2.8\%)    & 32.5 ($\pm$ 79.8)     & 149.2 ($\pm$ 96.6)&           \\ \midrule
- & Oracle  & 100\% ($\pm$ 0.0\%)    & 58.1 ($\pm$ 14.2)    & 80.8 ($\pm$ 25.1)              \\
- & Random  & 5.7\% ($\pm$ 2.6\%)    & -43.2 ($\pm$ 4.9)    & 248.9 ($\pm$ 15.1)             \\ \bottomrule
\end{tabular}}
\end{center}
\caption{
\textbf{PPO baseline performance on the Small dataset} for the door finding tasks. The maximum trajectory length was set to 253, the number of actions required by the oracle to achieve the longest task. All metrics from training 10 seeds (except for AllObs where we removed two outlier policies) for 2M frames, then averaging 1000 test episodes per seed. The standard deviation reported here is across seeds and evaluation episodes. The two bottom lines of the table corresponds to the mean optimal path length (as calculated by the oracle) and random mean reward for comparison.}
\label{tab:results}
\end{table}

As expected, the AllObs model which fuses image data, visible text, and GPS achieved the highest rate of successful navigations.
After 2M frames of experience, this model converged to a policy that can navigate to the 116 goals in Small with success achieved in $74.9\%$ of trajectories, with the best performing model achieving an $85\%$ success rate.
The performance of the AllObs model also improved much more quickly than any other model.
However, some trajectories still failed to achieve the goal resulting in a mean trajectory length nearly 40\% higher than the oracle. 
Upon inspection of the trained policies, we observed that the agent sometimes tried to make a forward action when blocked by a wall or street. The agent also sometimes repeated the same incorrect action, turning right and left.
Still, the agent very often exhibited "intelligent" behaviour, turning to follow the sidewalk and looking for house numbers.

\begin{figure}[htb]
\centering
\begin{subfigure}{.31\textwidth}
  \centering
  \includegraphics[width=\linewidth]{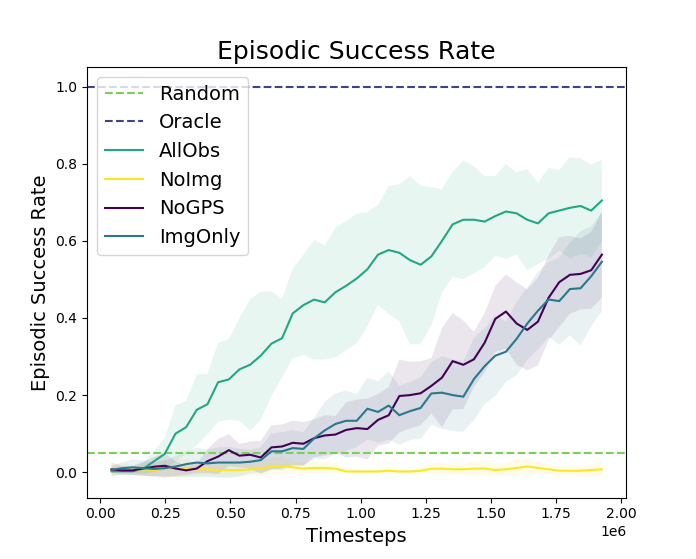}
  \caption{Success Rate}\label{fig:mini-shaped-success-rate}
\end{subfigure}%
\begin{subfigure}{.31\textwidth}
  \centering
  \includegraphics[width=\linewidth]{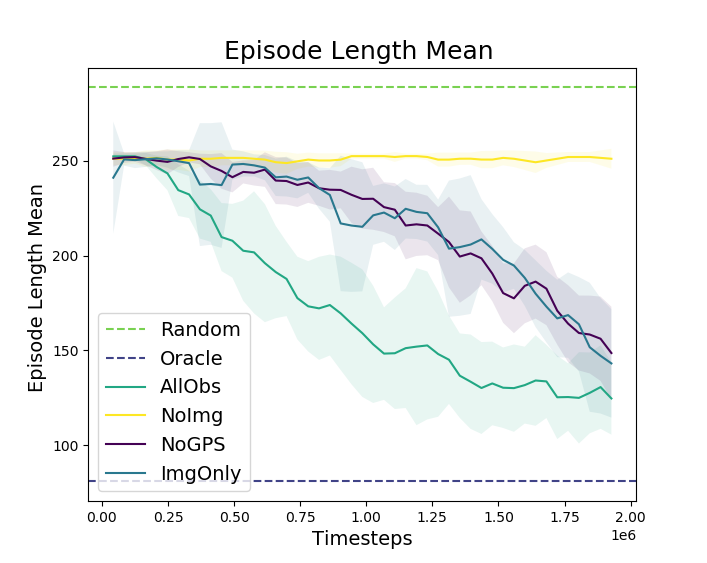}
  \caption{Trajectory length}\label{fig:mini-shaped-length}
\end{subfigure}%
\begin{subfigure}{.31\textwidth}
  \centering
  \includegraphics[width=\linewidth]{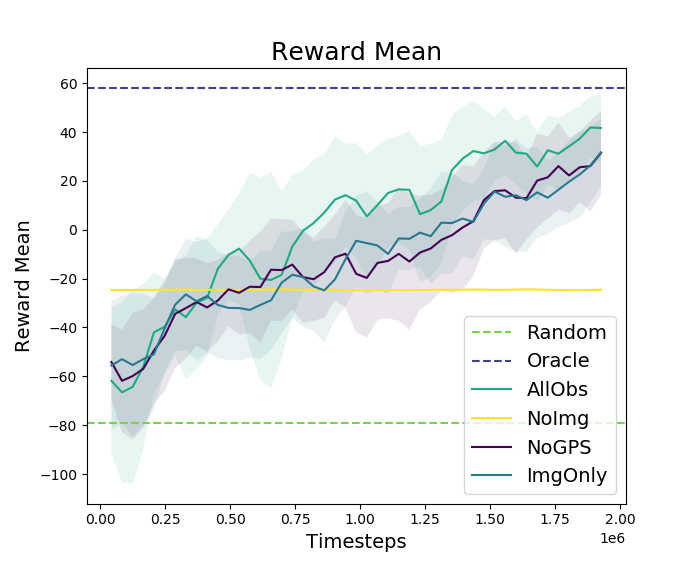}
  \caption{Trajectory rewards}\label{fig:mini-shaped-reward}
\end{subfigure}
\caption{\textbf{Training curves over 2M frames} of experience in the Small dataset. We report the mean and standard deviation over 10 seeds of several metrics smoothed with a moving average on 10 episodes. The oracle and random walk performances are also provided for comparison. (a) the proportion of episodes which terminate at the target door. (b) the length of an episode, which terminates after 253 actions or finding the target door. (c) the mean dense reward as described in subsection \ref{propreward}. 
}
\label{fig:mini-results}
\end{figure}

Every model with access to image data had above 55\% success rate on average, whereas the model without access to images (NoImg) learned a degenerate policy that performed even worse than a random walk.
The poor performance of the NoImg model indicates the importance of even down-scaled $84\times84$ pixel images for sidewalk navigation.
Judging from the moderately reduced performance of ImgOnly and NoGPS, the AllObs model seems not to rely solely on GPS or image data, but can use a combination of all input modalities to achieve superior performance.
The next best performing agent, ImgOnly, performed much better, completing the task in more than half of episodes. Adding the visible text modality (i.e., NoGPS model) seems to further improve performance when compared to the ImgOnly model. Figure \ref{fig:mini-results} shows the PPO training metrics for task 1 to 4 on Small.

\section{Limitations \& Conclusion}

The SEVN dataset and simulator is an \ac{RL} environment for sidewalk navigation. 
The codebase and dataset are open and extensible, providing a framework for future research into BVI navigation.
As with most machine learning models, deploying \ac{RL} models trained in this environment into the real-world presents significant challenges.
Techniques like domain randomization\cite{tobin2017domain} can be used to improve model generalization, but our use of real-world imagery presents a challenge for this approach.
One alternative to improve performance would be to gather additional imagery and annotations with better hardware and more sophisticated methods.
This could result in higher resolution, more precisely located images, with fewer stitching artifacts common in panoramic images. 
Alternative methods for embedding scene text as input for the policy model may also yield improved performance.

Notwithstanding this work's contributions, there remain many challenges ahead for those who wish to create assistive pedestrian navigation systems. 
One major challenge is to improve the efficiency of these systems, enabling deployment to edge devices while maintaining accuracy and speed.
Large-scale object detection, optical character recognition, and neural networks in general remain computationally intensive despite recent advances.
Another major challenge is to design a communication strategy for these assistive technologies that respects the user's level of vision, comfort, and safety.

\subsubsection*{Acknowledgments}
This research was supported by NSERC, PROMPT, CIFAR, MITACS, and ElementAI through both financial and computational resources. We would also like to thank Maryse Legault and François Boutrouille from HumanWare Technologies for sharing their expertise with developing tools for the blind and visually impaired. Further, we thank our colleagues at Mila, Yoshua Bengio, Hugo Larochelle, Myriam Côté, Liam Paull, Mandana Samiei, Jeremy Pinto, and Vincent Michalski, for engaging in discussions on computer vision, navigation, and reinforcement learning. Additionally, we would like to thank Megan Platts for reviewing and editing an earlier version of this manuscript. We are also grateful to the CoRL reviewers for their helpful feedback. 






\clearpage


\bibliography{corl}  

\clearpage
\section*{Supplementary materials}
\label{supplementary_material}

\section{Our User-centric Design Process}
\label{user-centric-design}

\subsection{Finding and Selecting a Problem}
Our design process began by conducting interviews with BVI people, experts in the field of accessibility, and developers of mobile apps for the BVI.  During this effort, we reviewed 50 products and services for the BVI community, which gave us some understanding of the market. 

We learned, with regards to navigation, that BVI people typically use white canes or guide dogs to avoid obstacles, and navigation tools like Google Maps and Blindsquare for route planning. For challenging visual tasks they use services like BeMyEyes or Aira.io which are human-in-the-loop systems that send images or videos to assistants who can cost over \$1/minute. We paid special attention to tasks which induced high willingness to pay for real-time human assistance because this indicates they are significant for the BVI person. In addition, we learned that many assistive technologies are made by individuals and relatively small companies, and though their attitudes varied, the developers were generally very open about sharing their views with us. The majority of the companies we emailed were focused on providing solutions to text summarization, vision, and navigation tasks.

In the end, our interviews and market research revealed three significant problems which are not yet fully resolved by available services or methods. These problems, namely intersection crossing, obstacle avoidance and finding specific locations, were constrained to sidewalk navigation. An example interview focused on navigation is contained in figure \ref{fig:interview}.

\textbf{Intersection crossing} is a very dangerous task for BVI people. To safely accomplish this task they must be aware of the spatial configuration of the intersection, the intersection's signalling pattern, the correct orientation to follow while crossing (Is there a crosswalk, underground passage, pedestrian bridge?), and the time when it is safe to begin crossing.

\textbf{Obstacle avoidance} happens when a person is walking on a sidewalk or crosswalk and is confronted by obstacles such as fire hydrants, parking meters, road work, stairs, etc. Today, this use case is primarily solved with traditional assistive tools, such as guide-dogs or canes. Other options like the BeAware app use a combination of beacons planted in the environment and bluetooth.

\textbf{Finding a specific location} happens when a person has a unique destination such as a restaurant, physician's office, or public transit station. The most difficult part of this task is the last few meters of navigation. BVI people often use smartphone apps like Google Maps that leverage GPS, but this only partially solves the problem. 
Once near the desired location, BVI people usually have to consult a scene description application or nearby people to make sure they are in the right place. However, objects of interest such as building entrances, ramps, stairs, house numbers, bus stops, and subway entrances are often missing.

After highlighting those problems, we decided to pursue the "finding a specific location" problem because other groups \citep{radwan2018multimodal, Diaz_2017_ICCV, DBLP:journals/corr/abs-1808-06887} had provided significant treatment to the "intersection crossing" problem, and the "obstacle detection" problem already has workable solutions in the form of white canes and guide dogs. Furthermore, the sidewalk navigation sub-problem seemed to have significant individual and commercial interest with limited prior work.

\begin{figure}[ht]
\framebox{
\begin{minipage}{\textwidth}
\vspace{-2mm}
\begin{multicols}{4}
\ssmall{
\hspace*{5pt}Maryse Legault is a visually impaired woman who works as an accessibility specialist at HumanWare, a global leader in assistive technology for people who are blind or have low vision needs. In this interview, we discussed her needs in order to feel comfortable travelling to new locations.

\hspace*{5pt}\textbf{Interviewer: Do you need to prepare for your journey?}

\hspace*{5pt}Maryse: Yes, like sighted people, I need to prepare myself for the trip. For example, I am more comfortable knowing which side of the street the metro station door or the bus stop post is on. If I need to cross a boulevard to get there, knowing if there is a crosswalk, traffic lights, an underground passage or a pedestrian bridge can be very helpful. I also need to know about the direction of traffic and potential obstacles like roadwork, especially if I know that I will not be able to reroute with a map on my phone.

\hspace*{5pt}\textbf{Interviewer: This seems to be a lot of preparation, but I hear that there are mobile phone applications that can be used to help with those difficulties on the ground. Can you give us examples of applications that you use?}

\hspace*{5pt}Maryse: That’s true. For example, when I take the bus I follow the Navigation app’s vocal instruction to know which street I am on. Sometimes it can be confusing, especially in the US where you can be heading East on West Robert Smith Avenue. This app helps me to get off at the correct stop. When I’m on a sidewalk, I must always ask other pedestrians question to figure out where I am. Like, what is the current house number? I can sometimes ask for help to cross intersections or even more. 

\hspace*{5pt}\textbf{Interviewer: Okay, so navigating the city seems to still be very challenging. What would be helpful for you to know to make navigation easier between intersections, and at the intersection?}

\hspace*{5pt}Maryse: Well, between intersections I would like to know about obstacles to avoid like orange cones and bicycles. At the intersection, knowing if there is a crosswalk, how many lanes I have to cross, if there are some traffic lights or not and if the light is green would be very helpful! There is an application that can help with general object detection but there is no feature that lets me select the kind of object I want to detect during navigation. I think I would like an app where I can select a mode which does not overwhelm me with too much information.

\hspace*{5pt}\textbf{Interviewer: It sounds like these apps give you more information than you need and are sometimes more annoying than helpful. I suppose that when you arrive at your destination you will be looking for certain types of objects. Can you tell me more about that?}

\hspace*{5pt}Maryse: First let’s give a bit of context. When I arrive at my destination I first need to check that I am at the right place. I usually do that by asking people around and the information I am asking for is very dependent on my destination. For example, if I am going to a shopping mall the main issue is that it’s typically an open space. These are more difficult to understand than streets and sidewalks. It is often difficult for me to walk between parking lots and pedestrian areas. There are very often few or no sidewalks from the bus stop to the shopping center door. So being able to detect those things would be very helpful. Knowing the store's location when arriving at the shopping center would also be a plus. Then I would also need information about what’s on the storefront like the store logo, the signs, the opening hours and the house number for example.

\hspace*{5pt}\textbf{Interviewer: Okay, and last question: When you are at the right location is the story over?}

\hspace*{5pt}Maryse: Well unfortunately not, indoor navigation is also very challenging. For example, if I am at a restaurant I need to know if there is any available seat at one of the tables and how to get to it. I might also need to know how to get to the counter. And I would say even before that finding the entrance or the door of the restaurant is very challenging as dogs are not really able to do that.}
\end{multicols}
\end{minipage}%
}
\captionof{figure}{\textbf{Interview example.} This type of interview was conducted with BVI people, orientation and mobility specialists, and developers of applications for the BVI.}
  \label{fig:interview}
\end{figure}

\subsection{Further Contextualizing the Sidewalk Navigation Problem}
Having narrowed our focus to a specific problem, sidewalk navigation to unique locations, we looked for relevant datasets and models available in the field of machine learning to solve this problem.

\textbf{Object detection.} Detecting objects like house numbers, doors and street name signs is necessary in order to complete the navigation task of finding the entrance to a specific address. A quick examination of the classes covered in general object detection benchmarks datasets like ImageNet \citep{ILSVRC15} reveal that they lack those object classes necessary for pedestrian navigation. Specialized datasets for autonomous vehicles like ApolloScape \citep{wang2019apolloscape} provide some of the classes of interest like sidewalks, bridges, walls, fences, traffic signs and signals. However, these classes were not comprehensive or fine-grained enough to achieve our tasks of interest and all the images are from the perspective of a motor vehicle, not a pedestrian. For house numbers, there is also the Street View House Number dataset, but the images are tightly cropped for image to text tasks instead of text detection \citep{netzer2011reading}. Therefore, we decided to label in SEVN some of the relevant object classes for sidewalk navigation that are missing in other datasets (doors, street name signs).

\textbf{Scene text recognition (STR).} Addresses usually contain a street name and house number that manifest as visible text in the environment. This text, as well as other scene text like business names, are useful landmarks for localization. Extracting incidental scene text can be quite difficult because it appears in the scene without the user having taken any specific prior action to cause its appearance or improve its positioning in the frame. Datasets of annotated images with real text \citep{wang2011end} and synthetically overlayed text \citep{jaderberg2014synthetic} have resulted in useful models, but modest end-to-end performance on the ICDAR 2019 Large-Scale Street View Text with Partial Labelling challenge indicates that this problem remains open \citep{DBLP:journals/corr/abs-1903-11800}. Unlike other datasets for STR, our dataset has high-resolution geo-localized images and can generate sequential views of an urban environment, creating an opportunity to investigate sequential text.

\section{Model Details}
\label{ppo-params}
\begin{table}[h]
\centering
\begin{tabular}{@{}ll@{}}
\toprule
\textbf{Hyperparameter}                       & \textbf{Value}         \\ \midrule
Learning Rate                                 & $3\times 10^{-4}$ \\
Number of Steps                               & 2048                    \\
Value Loss Coefficient                        & 0.5                    \\
Linear LR Decay Schedule                      & True                   \\
\midrule
Entropy Coefficient                           & 0                   \\
Gamma ($\gamma$)                                        & 0.99                   \\
Generalized Advantage Estimation ($\lambda$) & 0.95                   \\
Maximum Gradient Norm                         & 0.5                    \\
\midrule
Number of PPO Epochs                          & 4                      \\
Number of PPO Mini-Batches                    & 32                     \\
Clipping Parameter                            & 0.2                   \\
\bottomrule
\end{tabular}
\caption{
\textbf{Hyperparameters}. We report the hyper-parameters used while training. These settings are quite similar to those \ac{PPO} parameters described in \cite{DBLP:journals/corr/SchulmanWDRK17}.
} 
\end{table}

\section{Additional Maps}
The maps shown in the body of the work represent one view on our dataset. Here, we present several other visualizations of this geographic region.

\subsection{Goal Locations}
\begin{figure}[hb]
  \includegraphics[width=\linewidth]{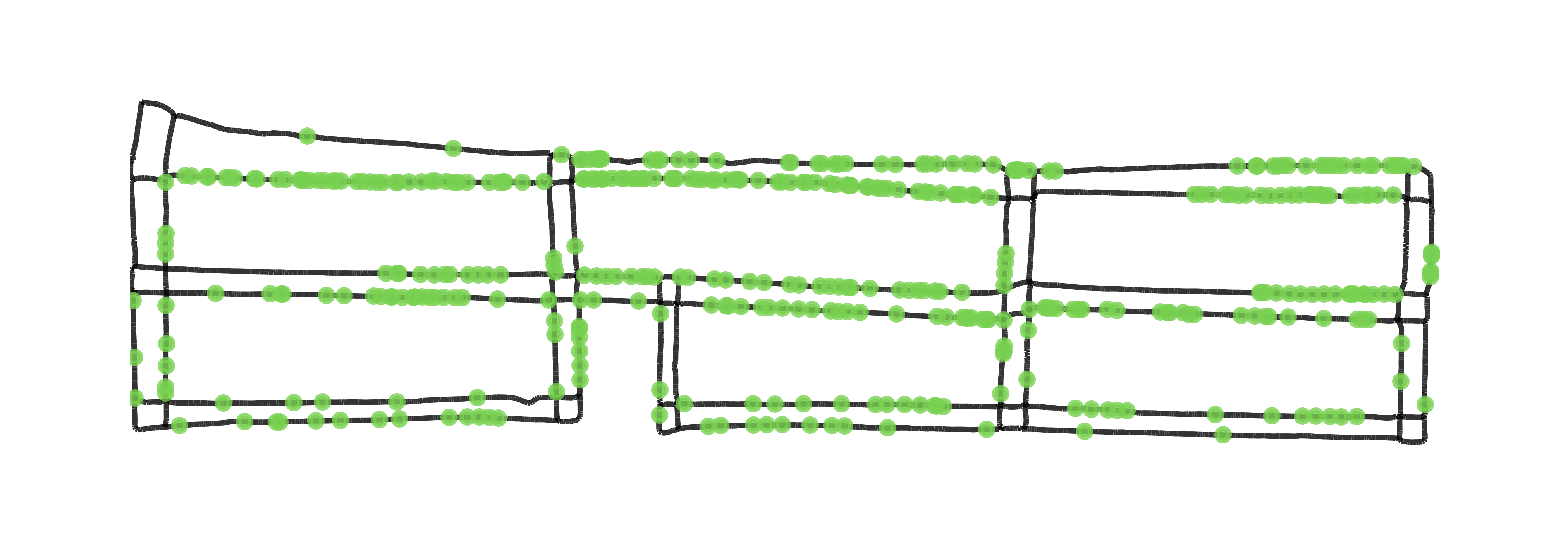}
  \captionof{figure}{\textbf{Map with Goal Locations} in the same orientation as previous maps, with the Small data split being the top right block. We can see that the target locations are well spread out, but still diverse, with some dense areas (often commercial or residential zones as seen in Figure \ref{fig:zones}), whereas parks and industrial zones are more sparsely populated with addresses.}
  \label{fig:goal-locs}
\end{figure}

In SEVN, doors with visually identifiable house numbers are available as goals. The distribution of these goals is dependent on many factors -- visibility of the house number, visibility of the door, clear relationship between the two, but most importantly the type and distribution of buildings along the street. In figure \ref{fig:goal-locs}, we show the spatial distribution of goals within SEVN.

\label{goal-locs}

\subsection{Zoning}
\label{zoning}
\begin{figure}[hb]
  \includegraphics[width=\linewidth]{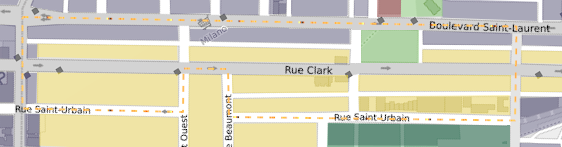}
  \captionof{figure}{
      \textbf{Usage and Zoning Map} for the region covered by SEVN, outlined by the \textbf{dotted orange} line. In this graphic, \textbf{purple} indicates commercial zones, \textbf{yellow} indicates residential zones, \textbf{green} indicates parks and outdoor recreation areas, \textbf{dark teal} is industrial, and \textbf{red} indicates collective or cultural institutions (in this case, a cathedral). Many areas are mixed, in the sense that they can be commercial and residential. These zones are colored based on their primary purpose. Zoning rules are complex and the details can be viewed here: \url{http://ville.montreal.qc.ca/pls/portal/docs/1/89510003.PDF}.  The basis of this map is the OpenStreetMap Carto (Black and White), with zoning data courtesy of the city of Montreal: \url{http://ville.montreal.qc.ca/pls/portal/docs/PAGE/ARROND_RPP_FR/MEDIA/DOCUMENTS/USAGES_PRESCRITS_P1_4.PDF}.
      }
  \label{fig:zones}
\end{figure}

Arguably the primary factor determining the distribution of goal locations is Montreal's urban zoning regulation. In figure \ref{fig:zones} we show the primary zone determined by the municipality of Montreal around and within the region captured by SEVN.

\section{Additional Experiments}
\subsection{Noisy GPS Ablation}
\label{noisy-gps}
\begin{figure}[htb]
\centering
\begin{subfigure}{.31\textwidth}
  \centering
  \includegraphics[width=\linewidth]{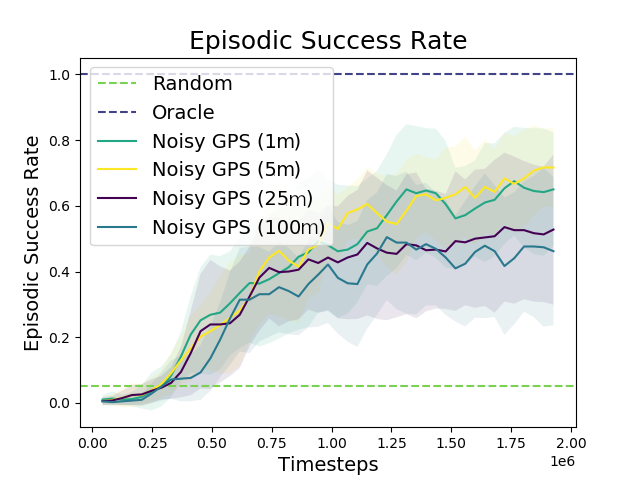}
  \caption{Success Rate}\label{figgps_exp_successes:}
\end{subfigure}%
\begin{subfigure}{.31\textwidth}
  \centering
  \includegraphics[width=\linewidth]{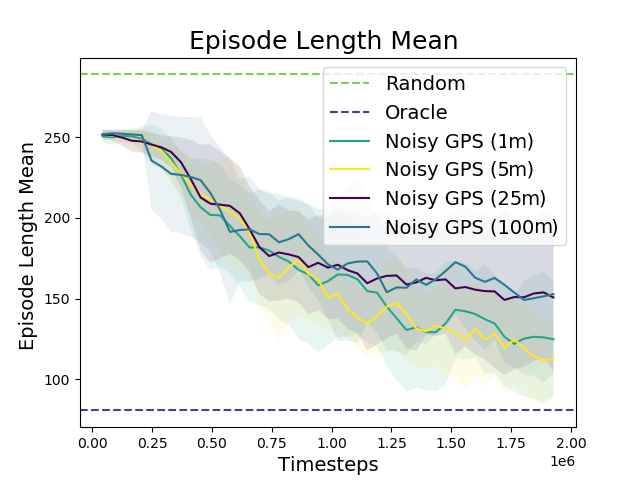}
  \caption{Trajectory length}\label{fig:gps_exp_ep_len}
\end{subfigure}%
\begin{subfigure}{.31\textwidth}
  \centering
  \includegraphics[width=\linewidth]{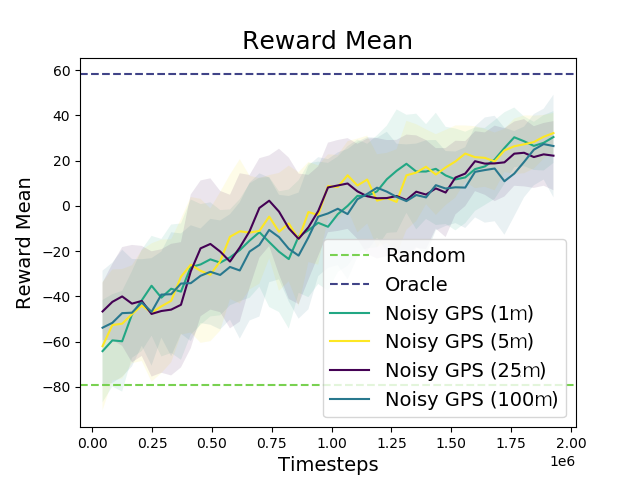}
  \caption{Trajectory rewards}\label{fig:gps_exp_reward}
\end{subfigure}
\caption{\textbf{GPS Ablation Study}. We report the mean and standard deviation over 10 seeds of several metrics smoothed with a moving average on 10 training episodes -- similar to that in Figure \ref{fig:mini-results}. (a) the proportion of episodes which terminate at the target door. (b) the length of an episode, which terminates after 253 actions or finding the target door. (c) the mean total trajectory reward. 
}
\label{fig:noisy-gps}
\end{figure}

Interested by the effect of noisy GPS sensors on navigation performance,
we created a setting where the agent had access to all types of observation modality, but we could add varying levels of Gaussian noise to the GPS sensor. To construct this setting, on the first timestep when we calculate the goal position by sampling the coordinates of a panorama which satisfies the goal conditions, we also sample a Gaussian noise ($\mu=0,\sigma=x$) where $x$ is the amount of noise in meters. 
Every time we compute a GPS-based goal offset, we take the agent's current position and add Gaussian Noise ($\mu=0,\sigma=x$) then subtract the noisy goal coordinates. Results shown in Figure \ref{fig:noisy-gps}.

These results, on the street segment task, indicate that GPS readings with large amounts of noise (25, 100) unsurprisingly lead to worse performance and higher variance in our model. This result is consistent with Figure \ref{fig:mini-results}, and provides evidence that a model with a highly noisy GPS sensor performs as well as a model trained without a GPS sensor (NoGPS). Smaller amounts of noise seem to have little impact on performance, and possibly even increase performance. However, the instability of \ac{DRL} algorithms can make it difficult to recognize small changes to performance.

\subsection{Generalization Experiment}
We hoped that our policy trained to perform navigation to goal doors would generalize to other regions. This, sadly, does not appear to be the case. However, the trained policies are still valuable within the training region. Further, we believe that given sufficient data these policies may generalize. Also, the model proposed in section \ref{sec:bas} was not specifically designed for generalization. In table \ref{tab:generalization} we present results of the fusion model with all modalities evaluated outside of the training data. 

\begin{table}[ht]
\begin{center}
\resizebox{0.66\textwidth}{!}{\begin{tabular}{@{}lll@{}}
\toprule
                       \multicolumn{1}{c}{\textbf{Evaluation}} & \multicolumn{1}{c}{\textbf{Trained on Large}} & \multicolumn{1}{c}{\textbf{Trained on Small}} \\ 
                       \midrule
Success Rate on Large   & 21.1\% ($\pm$ 3.7\%)          & 1.0\% ($\pm$ 0.3\%)          \\
Success Rate on Small    & 1.0\% ($\pm$ 0.4\%)           & 74.9\% ($\pm$ 7.8\%)          \\
Rewards on Large        & 13.3 ($\pm$ 70.4)          & -134.7 ($\pm$ 110.9)      \\
Rewards on Small         & -71.1 ($\pm$ 92.9)         & 43.9 ($\pm$ 65.9)        \\
Episode Length on Large & 213.5 ($\pm$ 78.9)        & 250.5 ($\pm$ 24.5)        \\
Episode Length on Small  & 250.6 ($\pm$ -24.2)         & 113.7 ($\pm$ 89.7)       \\
\bottomrule
\end{tabular}}
\end{center}
\caption{\textbf{Generalization Results}: This is an extension of table \ref{tab:results}, with the fusion model trained on the Small dataset remaining the same. We also evaluated this model on the Large set (all data except Small). We also evaluated the same model trained for 20M frames of experience on Large.}
\label{tab:generalization}
\end{table}

We can see that the results of the model trained on the Small dataset and evaluated on the Small dataset performed well, with the same 74.9\% success rate seen in table \ref{tab:results}. When evaluated on the rest of the dataset, this model failed with a success rate of only 1.4\%. Similarly poor performances are seen for the model trained on the (much larger) Large dataset with 20M frames of experience. 

\end{document}